%% file: main.tex
\documentclass{article}
\usepackage{ijcai26}

\usepackage{times}
\usepackage{soul}
\usepackage{url}
\usepackage[hidelinks]{hyperref}
\usepackage[utf8]{inputenc}
\usepackage[small]{caption}
\usepackage{graphicx}
\usepackage{amsmath}
\usepackage{amsthm}
\usepackage{amssymb}
\usepackage{booktabs}
\usepackage{algorithm}
\usepackage{algorithmic}
\usepackage[switch]{lineno}
\usepackage{subcaption}
\usepackage[]{algorithm,array}
\usepackage{ifthen}
\usepackage{pifont}
\usepackage{pgfplots}
\usepackage{listings}
\usepackage{float}

\usepackage{multirow}
\newtheorem{problem}{Problem}
\DeclareMathOperator*{\argmax}{argmax}

\pgfplotsset{compat=1.18}

\title{Label Curation Using Agentic AI}

\author{
Subhodeep Ghosh
\and
Bayan Divaaniaazar\and
Md Ishat-E-Rabban\and
Spencer Clarke\And
Senjuti Basu Roy\\
\affiliations
New Jersey Institute of Technology\\
\emails
\{sg2646, bd347, mi375, sdc2, senjutib\}@njit.edu
}

\begin{document}

\maketitle

\begin{abstract}
    
Data annotation is essential for supervised learning, yet producing accurate, unbiased, and scalable labels remains challenging as datasets grow in size and modality. Traditional human-centric pipelines are costly, slow, and prone to annotator variability, motivating reliability-aware automated annotation. We present {\tt AURA} ({\bf A}gentic AI for {\bf U}nified {\bf R}eliability Modeling and {\bf A}nnotation Aggregation), an agentic AI framework for large-scale, multi-modal data annotation. {\tt AURA} coordinates multiple AI agents to generate and validate labels without requiring ground truth. At its core, {\tt AURA} adapts a classical probabilistic model that jointly infers latent true labels and annotator reliability via confusion matrices, using Expectation-Maximization to reconcile conflicting annotations and aggregate noisy predictions. Across the four benchmark datasets evaluated, {\tt AURA} achieves accuracy improvements of up to 5.8\% over baseline. In more challenging settings with poor quality annotators, the improvement is up to 50\% over baseline. {\tt AURA} also accurately estimates the reliability of annotators, allowing assessment of annotator quality even without any pre-validation steps. 
\end{abstract}

\input{intro}
\input{datamodel}

\input{method}
\input{exp}

\input{related}
\input{conc}

\bibliographystyle{named}
\bibliography{reference}

\end{document}

%% file: intro.tex
\section{Introduction}
Data annotation~\cite{huang2024data,yan2018cost,zhdanovskaya2023data,pan2025enhanced} is a fundamental prerequisite for supervised learning, yet producing accurate, unbiased, and scalable labels remains a persistent challenge as datasets grow in size, diversity, and modality. Traditional human-centric annotation pipelines rely heavily on expert or crowd-sourced labeling~\cite{crowd1,crowd2,crowd3}, which is costly, time-consuming, and difficult to scale. Moreover, human annotations often exhibit significant variability due to subjective interpretation, fatigue, or uneven expertise, leading to noisy and unreliable labels.

Active learning partially alleviates this burden by judiciously involving human annotators in the labeling loop~\cite{basu2019capturing,wang2017active,yan2018cost}. However, despite reducing the number of required annotations, active learning remains fundamentally constrained by human availability, latency, and cost, rendering it insufficient for modern large-scale, continuous, and multi-modal data streams.

Recent advances in large language models (LLMs) and autonomous agents have spurred a growing body of work on fully automated annotation pipelines, where multiple AI agents collaboratively generate, refine, and validate labels~\cite{agentic1,agentic2,agentic3,agentic4,agentic5,agentic6,agentic7,agentic8}. Although these approaches highlight the promise of agent-driven automation, existing work largely overlooks the systematic modeling of uncertainty, noise, and bias inherent in such agents. In practice, autonomous agents are imperfect and heterogeneous in reliability, with performance varying across tasks, modalities, and contexts. This exposes a critical need for a statistically grounded and generic annotation framework that explicitly model agent uncertainty, disagreement, and bias, and infer high-confidence labels in the absence of ground truth to any domain without additional fine-tuning or in-context learning. Equally important is the ability to rigorously quantify and track the accuracy and reliability of individual agents. Designing such reliability-aware, agentic annotation systems is therefore essential for enabling scalable data annotation for next-generation AI applications.

We present {\tt AURA} (\textbf{A}gentic AI for \textbf{U}nified \textbf{R}eliability Modeling and \textbf{A}nnotation Aggregation), a fully automated agentic AI framework for large-scale, multi-modal data annotation. {\tt AURA} coordinates multiple AI agents - each exhibiting distinct strengths, and error characteristics - to collaboratively generate, refine, and aggregate annotations. {\em Importantly, {\tt AURA} operates entirely with off-the-shelf AI models and does not require any model fine-tuning or in-context learning, allowing it to generalize seamlessly across tasks, modalities, and domains.} Rather than treating annotations as independent votes, {\tt AURA} explicitly models annotator reliability through a statistically grounded probabilistic framework, enabling principled aggregation of conflicting predictions and the inference of high-confidence labels in the absence of gold-standard supervision.

A core innovation of {\tt AURA} is the Algorithm {\tt AEML} (Agentic Expectation Maximization Labeling) that is designed by appropriately adapting a classical probabilistic model~\cite{dawid1979maximum}. This model is typically used for crowdsourced label aggregation that jointly infers latent true labels and annotator reliabilities through confusion matrices. {\tt AEML} resolves the inherent circular dependency between label correctness and annotator trustworthiness using an Expectation-Maximization (EM) technique~\cite{EM}. In the E-step, it estimates posterior distributions over the true labels by weighting each agent's predictions according to its current reliability estimates. In the M-step, these posterior label estimates are used to update each agent's confusion matrix via expectation-weighted counts of correct and incorrect annotations. The process repeats until convergence.

We conduct extensive experiments on four real-world image and video datasets to evaluate the effectiveness of {\tt AURA}. The results consistently show that annotations aggregated by {\tt AURA},  outperform annotations produced by individual annotators across all datasets, achieving higher performance. Moreover, the annotator reliability estimates learned by {\tt AURA} closely track the true annotator accuracies in most cases, demonstrating that the framework automatically identifies which agents are more reliable overall and which agents are better suited for specific classes. When compared against baselines, {\tt AURA} consistently yields superior performance, particularly in settings with heterogeneous annotator quality or class-dependent biases. {\tt AURA}'s consistent performance across diverse annotator architectures including large language models (GPT, Gemini), specialized vision models (Swin, CLIP), and domain-specific systems (Pegasus-1.2, QwenVL), demonstrate its adaptability and independence from task-specific calibration or architectural assumptions.

The rest of the paper is organized as follows - in Section~\ref{Sec:datamodel}, we present our data model and formalize the problems. Section~\ref{Sec:aura} presents the proposed framework and the algorithms inside it. In Section~\ref{sec:exp}, we present the experimental evaluation. Section~\ref{sec:related} presents related work and we conclude in Section~\ref{sec:conc}.

%% file: datamodel.tex
\section{Data Model \& Problem Definition}\label{Sec:datamodel}

\noindent {\bf Items and labels.}
Let $\mathcal{X} = \{x_1, x_2, \ldots, x_n\}$ denote a set of $n$ data instances and let $\mathcal{Y}$ denote the label space. 

\noindent {\bf Annotators and Annotator Reliability.}
$\mathcal{A} = \{1,\ldots,M\}$ be a set of annotators. An \emph{annotator} $a \in \mathcal{A}$ is an agent that assigns a label $\ell_i^{(a)} \in \mathcal{Y}$ to an instance $x_i \in \mathcal{X}$. $\ell^{(a)}$ represents a label assigned by annotator $a$. 

\noindent {\bf Ground Truth.}
Let $y_i^\star$ denote the (possibly latent) ground-truth label associated with instance $x_i$. In many real-world annotation tasks, $y_i^\star$ is not directly observable and must be inferred from multiple noisy annotations or expert adjudication. $y^\star \in \mathcal{Y}$ represents a latent ground-truth label.

\noindent {\bf Class Priors.} Also known as prior probability, $p_{y^\star}$ defines how frequently label $y^\star$ appear in the dataset and is given by:
\begin{equation} \label{eq1}
p_{y^\star}= \frac{1}{n}\sum_{i=1}^n\Pr\!\left(y_i^\star = y^\star\right).
\end{equation}

\noindent{\bf Annotator Reliability.}
The reliability of annotator $a$ is defined as the probability that the annotator assigns the correct label:
\begin{equation} \label{eq2}
\alpha_a \;=\; \Pr\!\left(\ell^{(a)} = y^\star\right).
\end{equation} \label{eq:prob_est_correct}
Given a set $\mathcal{X}_a \subseteq \mathcal{X}$ of instances annotated by $a$, it is estimated as:
\begin{equation} \label{eq3}
\hat{\alpha}_a
\;=\;
\frac{1}{|\mathcal{X}_a|}
\sum_{x_i \in \mathcal{X}_a}
\mathbb{I}\!\left[\ell_i^{(a)} = y_i^\star\right],
\end{equation}
where $\mathbb{I}[\cdot]$ denotes the indicator function.

\noindent {\bf Convergence Threshold}. Let $\gamma$ denote the convergence threshold. If the difference in log-likelihood between two consecutive iterations fall below $\gamma$ then we assume the algorithm has converged.

\begin{problem}[Ground-Truth Label Identification]
Let $\mathcal{X} = \{x_1,\ldots,x_n\}$ be a set of $n$ instances, each associated with a latent ground-truth label $y_i^\star \in \mathcal{Y}$.
Let $\mathcal{A} = \{1,\ldots,M\}$ denote a set of annotators who provide noisy
labels $\ell_i^{(a)} \in \mathcal{Y}$ associated with the instances.

Given the observed annotations $\{\ell_i^{1:M}\}$ and assuming that annotators may
be unreliable and biased, the objective is to infer, for each instance $x_i$, its
latent ground-truth label $y_i^\star$ (or equivalently, posterior probability of $y^\star$ with respect to $x_i$ as 
$w_{iy^\star}\! =\! \Pr(y_i^\star\! =\! y^\star \mid \{\ell_i^{1:M}\}, p_{y^\star}, \{\Theta^{1:M}\})$) that best explains the observed annotations.
\end{problem}

\begin{problem}[Annotator Reliability Matrix Estimation]
Let $\mathcal{A}\! =\! \{1,\ldots,M\}$ be a set of annotators, each characterized by an
unknown reliability profile represented as a confusion matrix
$\Theta^{(a)}\! \in\! [0,1]^{|\mathcal{Y}| \times |\mathcal{Y}|}$, where
$\Theta^{(a)}_{k,l}\!\! =\!\! \Pr(\ell^{(a)}\! =\! l \!\mid\! y^\star\!\! =\!\! k)  \forall k,l \in \mathcal{Y}$ ($k,l$ together represent an index in $\Theta^{(a)}$). 

Given a collection of instances with latent ground-truth labels and the observed annotations
$\{\ell^{1:M}\}$, the objective is to estimate the annotator reliability matrices
$\{\Theta^{1:M}\}$ that best explain the observed labeling behavior, accounting for
the uncertainty in the latent ground-truth labels.
\end{problem}

%% file: method.tex
\section{Proposed Solution Framework}\label{Sec:aura}
\begin{figure}[htbp]
    \centering
    \includegraphics[width=\linewidth, height=7cm]{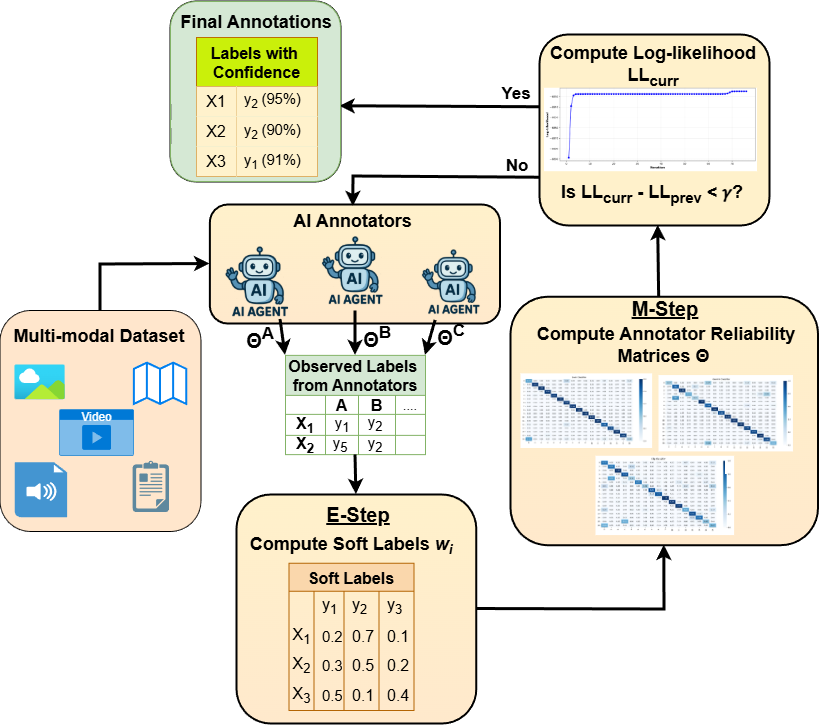}
    \caption{{\tt AURA} framework}
    \label{fig:framework}
\vspace{-0.1in}
\end{figure}

Figure~\ref{fig:framework} presents {\tt AURA}, an Agentic AI based solution framework, which takes off-the-shelf AI agents without any fine tuning or in-context learning. Each agent acts as an independent annotator with its own (unknown) accuracy profile.  Given a dataset with no available ground-truth labels, instances are processed in parallel by multiple agents, each of
which analyzes the input and produces a class prediction according to a
model-specific prompting and inference pipeline.

 In this setting, AI agents (i.e., annotators) are inherently heterogeneous. Consequently, simply treating all agents as equally reliable-an assumption implicitly made by standard ensemble techniques such as majority voting-can lead to suboptimal and unstable predictions. While prior work typically addresses ground-truth label inference and annotator reliability estimation as separate problems, our framework explicitly models their interdependence.  We design a probabilistic aggregation model to jointly infer the latent true label of each instance and the reliability of each agent.  By iteratively updating posterior label distributions and agent-specific confusion matrices, the framework enables principled aggregation of predictions in fully unlabeled, multi-modal settings.

\subsection{Algorithms Inside {\tt AURA}}
Now, we describe an expectation-maximization~\cite{EM} based algorithm {\tt AEML} (Agentic EM for Labeling) (refer to Algorithm~\ref{alg:ds}) that adapts the classical Dawid-Skene~\cite{dawid1979maximum} model that jointly estimates the annotators reliability and aggregated label of each instance.

 {\tt AEML} aggregates noisy labels from multiple annotators when ground-truth labels are unobserved. It assumes that each data instance has a latent true class and that annotators generate labels independently conditioned on this true class. Each annotator is characterized by a confusion matrix that encodes their reliability and systematic biases across classes, allowing the model to capture heterogeneous annotator behavior rather than assuming uniform accuracy.

Given an instance $x_i$, {\tt AEML} jointly estimates it's {\bf E}xpected true label $w_{iy^\star}\! =\! \Pr(y_i^\star\! =\! y^\star \mid \{\ell_i^{1:M}\}, p_{y^\star}, \{\Theta^{1:M}\}$ using class prior $p_{y^\star}$, responses $\ell_{i}^{(a)} \in \mathcal{Y}$ for every annotator $a \in \mathcal{A}$ and {\bf M}aximize the estimated likelihood of annotator reliability matrix $\Theta^{(a)} = \Pr(\ell^{(a)} | y^\star)$ . We iterate these two steps until the model converges, i.e., the difference in log-likelihood of consecutive iterations falls below a threshold value. In the E-step, the model computes posterior probabilities over the latent true label of each instance by combining the observed annotations weighted by the current estimates of annotator reliability matrices and class priors. In the M-step, these parameters are updated by maximizing the expected complete-data log-likelihood given the posteriors. Each EM iteration is guaranteed to monotonically increase the data likelihood, and the algorithm converges to a stationary point.

 \begin{algorithm} [!htbp]
\caption{Algorithm {\tt AEML}}\label{alg:ds}
\begin{algorithmic}[1]
    \STATE \textbf{Input:} $ \{1:\{x_1:\ell_1^{(1)}, 2: \ell_2^{(1)}, \dots, x_n: \ell_n^{(1)}\},\dots M:\{\dots \}\}$  \label{algline: 1}
    
    \STATE \textbf{Output:} $\left<x_i, \argmax_{y^\star}({w_{iy^\star}}), w_{iy^\star}\right>,\forall x_i\in \mathcal{X}$  \label{algline: 2}
    
     \STATE \textbf{Initialize:} $p_{y^\star} \!\gets\! \frac{1}{|\mathcal{Y}|} \forall y^\star\! \in\! \mathcal{Y}$, $\Theta_{k = l}^a \!\gets\! \frac{\lambda}{100}$, $\Theta_{k \neq l}^a \!\gets\! \frac{100 - \lambda}{100(|\mathcal{Y}|-1)} \forall a\! \in\! \mathcal{A}\!  =\! \{1,\ldots,M\}$, 
    $LL_{curr} \!\gets\! 0$, $LL_{prev} \!\gets\! -\infty$    \label{algline: 3}
    \WHILE{$True$}                                             \label{algline: 4}
        \FOR{each $x_i \in \mathcal{X}$}                        \label{algline: 5}
            \STATE E-step using equation~\ref{eq4}              \label{algline: 6}
        \ENDFOR                                                 \label{algline: 7}
        \FOR{each annotator $a \in \mathcal{A}$}                \label{algline: 8}
            \STATE M-step using equations~\ref{eq5} and \ref{eq6}   \label{algline: 9}
        \ENDFOR                                                   \label{algline: 10}
        \STATE Compute $LL_{curr}$ using equation~\ref{eq7}         \label{algline: 11}
        \IF{$LL_{curr} - LL_{prev}<\gamma$}                         \label{algline: 12}
            \STATE $break$                                          \label{algline: 13}
        \ELSE                                                      \label{algline: 14}
            \STATE $LL_{curr} \gets LL_{curr}$                          \label{algline: 15}
        \ENDIF                                                     \label{algline: 16}
    \ENDWHILE                                                      \label{algline: 17}
\end{algorithmic}
\end{algorithm}

\noindent {\bf Line~\ref{algline: 1}:} {\tt AEML} receives a dictionary of observed annotator labels $\ell_i^{(a)}$ corresponding to each instance $x_i \in \mathcal{X}$, for every annotator $a \in \mathcal{A}$.

\noindent {\bf Line~\ref{algline: 2}:} The label with the highest expectation for a particular instance is chosen as the predicted label for that instance with a confidence score equal to the expectation itself.

\noindent {\bf Line~\ref{algline: 3}:} Initially it is assumed every class has equal likelihood to appear and hence class priors $p_{y^\star}$ are initialized with uniform probability distribution. It also assumes that every classifier is $\lambda\%$ reliable at predicting the true label, i.e, we initialize the diagonal of $\Theta^a$ with $\frac{\lambda}{100}$, and distribute the remaining $(1-\frac{\lambda}{100})$ uniformly across $|\mathcal{Y}|-1$ classes.  $\lambda$ is a hyperparameter. The current and previous log-likelihood values are initialized with $0$ (best case) and $-\infty$ (worst case) respectively. 

\noindent {\bf Lines~\ref{algline: 4}-~\ref{algline: 17}:} {\tt AEML} learns the label for each instance and the reliability of each annotator by jointly iterating the EM-steps until convergence is reached.
 
\noindent {\bf Lines~\ref{algline: 5}-~\ref{algline: 7}:} In the E-step, the expected label of every instance is calculated using the equation:

\begin{align}\label{eq4}
    w_{iy^\star} &= \Pr(y_i^\star=y^\star \mid \{\ell_{i}^{1:M}\},p_{y^\star},\{\Theta^{1:M}\}) \\
    &= \frac{p_{y^\star}\prod_{a=1}^M \Pr\left(\ell_{i}^{(a)} \mid y_i^\star=y^\star, \Theta^a\right)}{\sum_{y^\star \in \mathcal{Y}} p_{y^\star}\prod_{a=1}^M \Pr\left(\ell_{i}^{(a)} \mid y_i^\star=y^\star, \Theta^a\right)} \nonumber
\end{align}

\noindent Equation~\ref{eq4} gives us the probability of instance $x_i$ belonging to class $y^\star\! \in\! \mathcal{Y}$, also called the {\bf posterior probability} of $y^\star$.
$\ \Pr\!\left(\ell_{i}^{(a)}\! \mid y_i^\star\!=\!y^\star, \Theta^a\!\right)$ can be obtained exactly from $\Theta_{y^\star, \ell_i^{(a)}}^a$.

\noindent {\bf Lines~\ref{algline: 8}-~\ref{algline: 10}:} The M-step maximizes the estimated likelihood of the annotator reliability matrix for every annotator $a \in \mathcal{A}$ by weighting the frequency of each observed label $\ell_i^{(a)}$ by the posterior probability $ w_{iy^\star}$ of the ground-truth class:

\begin{align}\label{eq5}
    \Theta^{(a)}_{k,l}\!\! &=\!\! \Pr(\ell^{(a)}\! =\! l \!\mid\! y^\star\!\! =\!\! k)  \\
    &= \frac{\sum_{i=1}^n w_{ik}.\mathbb{I}\!\left[\ell_i^{(a)} = l\right]}{\sum_{i=1}^n w_{ik}} \nonumber
\end{align} 

\noindent The denominator subjects the maximization to the constraint 
$
    \sum_{l=1}^{|\mathcal{Y}|} \Theta^{(a)}_{k,l} = 1
$
for given annotator $a$ forming a valid conditional probability distribution over all possible observed labels $l \in \mathcal{Y}$ given ground-truth class $k\in \mathcal{Y}$, keeping $\Theta^{(a)}$ mathematically interpretable and the model probabilistically coherent. The M-step effectively allows the framework to learn how reliable each annotator is in correctly identifying an instance from a particular class. 
The class priors are also updated as follows:
\begin{equation} \label{eq6}
    p_{y^\star} = \frac{1}{n}\sum_{i=1}^n w_{iy^\star}
\end{equation}

\noindent {\bf Line~\ref{algline: 11}:} The log-likelihood of current iteration is computed:
\begin{align}\label{eq7}
    LL_{curr} = \sum_{i=1}^n \log \left( \sum_{y^\star \in \mathcal{Y}} p_{y^\star} \prod_{a=1}^{M} \Theta^a_{y^\star,\ell_i^{(a)}} \right)
\end{align} 
Monitoring log-likelihood provides a quantitative measure of model fit. 

\noindent {\bf Lines~\ref{algline: 12}-~\ref{algline: 16}:} We check if the difference $LL_{curr} - LL_{prev} < \gamma$. This serves as a critical stopping criterion for {\tt AEML}, ensuring model parameters have stabilized and further iterations yield negligible improvement. Otherwise, we assign $LL_{prev}$ to $LL_{curr}$ and begin the next iteration.

\noindent {\bf Running time of {\tt AEML}.}
The E-step involves finding the product of the observed levels for all $M$ annotators with respect to each $|\mathcal{Y}|$ labels taking $\mathcal{O}(M \times |\mathcal{Y}|)$. This is repeated for all $n$ instances taking a total of $\mathcal{O}(n \times M \times |\mathcal{Y}|)$.
The M-step requires the computation of the annotator reliability matrix $\Theta^a$ which is of size $|\mathcal{Y}| \times |\mathcal{Y}|$ for all $M$ annotators. For each value in $\Theta^a$, we also need to calculate the summation of observed labels over $n$ instances. Thus the total time required is of the order $\mathcal{O}(n \times M \times |\mathcal{Y}|^2)$. 
The log-likelihood computation involves the summation of $n$ instances over the summation of $|\mathcal{Y}|$ classes with respect to the product of $M$ annotators resulting in a complexity of $\mathcal{O}(n \times M \times |\mathcal{Y}|)$.
Given the algorithm runs for $e$ epochs, the total time complexity of {\tt AEML} is $\mathcal{O}(e \times n \times M \times |\mathcal{Y}|^2)$.

%% file: exp.tex
\section{Experimental Evaluations}\label{sec:exp}

All experiments are conducted on an HP-Omen 16-n0xxx (Windows 11 Home v10.0.26100, Python 3.9.13) with an 8-core AMD Ryzen 7 6800H CPU and 32 GB RAM. 

\noindent Code is made available at:~\cite{Git}

\subsection{Experimental Setup}
We use the following datasets. The associated labels of the items are considered ground-truth for evaluation purposes.
\subsubsection{Datasets}

\begin{table}[h!]
    \centering
    \begin{tabular}{|l|c|r|r|}\toprule
        \textbf{Dataset name}  & \textbf{Modality} & \textbf{Total records} & \textbf{Total labels} \\ \midrule
        Kinetics-400& Video& 1000& 60\\
        \hline
          ImageNet-ReaL & Image & 4,271 & 50 \\
        \hline
          Food-101 & Image & 1,000& 101\\
        \hline
         CUB-200 & Image & 1018 & 17 \\ \bottomrule
    \end{tabular}
    \caption{Summary of  datasets}
    \label{tab:datasets}
 \vspace{-0.2in}
\end{table}

\noindent \textbf{Kinetics-400}~\cite{kay2017kineticshumanactionvideo} is an action recognition dataset that comprises more than 300,000 ten-second videos gathered from YouTube and annotated by human experts within 400 action categories that include sports, daily activities, and social behavior. 
For experimentation, 1,000 videos from 60 different classes were non-uniformly sampled.
\noindent \newline \textbf{ImageNet-ReaL} comprises of ImageNet~\cite{ILSVRC15} validation images augmented with a set of "Re-Assessed" (ReaL) labels~\cite{beyer2020imagenet}. For experimentation, 4271 images with multiple labels were assigned to the 50 most popular classes, distributed non-uniformly.
\noindent \newline \textbf{Food-101}~\cite{bossard14} is a food recognition dataset that comprises 101,000 images and classified within 101 food categories spread uniformly. For experimentation, 1000 images uniformly spanning all 101 classes were selected.
\noindent \newline \textbf{CUB-200}~\cite{wah_branson_welinder_perona_belongie_2011}, is an image classification benchmark containing 11788 images from 200 bird species with high visual similarity. For experimentation, 1018 images spread uniformly across 17 classes were selected. 

\begin{table*}[!htbp]
\centering
\begin{tabular}{l cc cc cc cc}
\toprule
\multirow{2}{*}{\textbf{Dataset}} 
& \multicolumn{2}{c}{\textbf{Accuracy}} 
& \multicolumn{2}{c}{\textbf{Precision}} 
& \multicolumn{2}{c}{\textbf{Recall}} 
& \multicolumn{2}{c}{\textbf{F1-score}} \\
\cmidrule(lr){2-3}\cmidrule(lr){4-5}\cmidrule(lr){6-7}\cmidrule(lr){8-9}
& {\tt AURA} & {\tt Maj.} 
& {\tt AURA} & {\tt Maj.} 
& {\tt AURA} & {\tt Maj.} 
& {\tt AURA} & {\tt Maj.} \\
\midrule
Kinetics-400   & 0.938 & 0.924 & 0.931 & 0.944 & 0.937 & 0.921 & 0.928 & 0.926 \\
ImageNet-ReaL & 0.831 & 0.786 & 0.841 & 0.847 & 0.828 & 0.799 & 0.834 & 0.822 \\
Food-101      & 0.827 & 0.769 & 0.846 & 0.886 & 0.824 & 0.760 & 0.818 & 0.791 \\
CUB-200       & 0.915 & 0.858 & 0.919 & 0.869 & 0.916 & 0.815 & 0.915 & 0.826 \\
\bottomrule
\end{tabular}
\caption{Comparison of {\tt AURA} and Majority Voting across Accuracy, Precision, Recall, and F1-score}
\label{tab:aura_all_metrics}
\end{table*}

\subsubsection{Agentic AI Models.}
The models annotate independently to prevent any form of inter-annotator influence or bias propagation. No pre-training or fine-tuning was involved. The models were directly used for inference. Images or videos were provided to each model by means of a prompt instructing the model to annotate items into one of the given classes based on the dataset as follows. 

\noindent {\bf Models Annotating Kinetics-400.}
The following $7$ models were used to annotate the videos via API calls: Google's \emph{Gemini-2.5-flash}, Alibaba Model Studio's \emph{Qwen-2.5-VL}, TwelveLab's \emph{Pegasus-1.2} - models with native video support - and, OpenAI's \emph{GPT-4o-mini, GPT-5-mini}, Replicate's \emph{LLaVA-13b, MoonDream}2 - which needed a preprocessing step where image frames were extracted from the respective videos using OpenCV~\cite{opencv_library}. The frames were then sent sequentially to the models for annotation. 

\noindent {\bf Models Annotating ImageNet-ReaL.}
The following $4$ models were used for image annotation via API calls: OpenAI's {\em GPT-4.1 Nano}, {\em GPT-4.1 Mini}, {\em GPT-4o}, and xAI's {\em Grok-2-vision-1212}.

\noindent {\bf Models Annotating Food-101.}
The following $4$ models were used for image annotation via 
local inference: Alibaba Model Studio's {\em Qwen3-vl-30b, Stella-vlm-32b, Internvl3\_5-30b-a3b}, and {\em Nemesis}.
Quantized versions of these models, provided by LMStudio\cite{lmstudio}, were used to enable efficient inference on local hardware.

\noindent {\bf Models Annotating CUB-200.}
The following $3$ models were used for image annotation via API calls: {\em Swin Transformer}~\cite{liu2021Swin}, Google's \emph{Gemini-2.5-flash}, and {\em CLIP Transformer}~\cite{radford2021learningtransferablevisualmodels}.
A synthetic {\em Random} annotator which matches the ground-truth with a fixed probability of $60\%$ was also included to study the effect of random noise and systematic errors in our framework. 

\subsubsection{Implemented Baseline} 

We use Majority Voting as a baseline for evaluating {\tt AURA}. Other related approaches are not directly applicable in our setting, as they require additional domain-specific fine-tuning or in-context learning, whereas {\tt AURA} operates using off-the-shelf models without task-specific adaptation.

\subsubsection{Evaluation measures} We use the following metrics for quantitative evaluation:
\noindent \newline \textbf{Accuracy} denotes the overall proportion of correctly classified instances across all classes
\noindent \newline \textbf{Precision} quantifies the proportion of predicted positive instances that are correctly classified, reflecting model's ability to avoid false positives
\noindent \newline \textbf{Recall} measures the sensitivity to detecting all relevant cases, that is the proportion of actual positive instances that are correctly identified
\noindent \newline \textbf{F1-score} represents the harmonic mean of precision and recall, providing a balanced metric that accounts for both false positives and false negatives - important in scenarios with class imbalances.
\noindent \newline \textbf{Support} is the number of samples present in each class within the dataset.
\noindent \newline \textbf{Pearson Correlation Coefficient} measures the linear relationship between two variables.

\begin{figure}[htbp]
    \centering
    \includegraphics[width=0.9\linewidth]{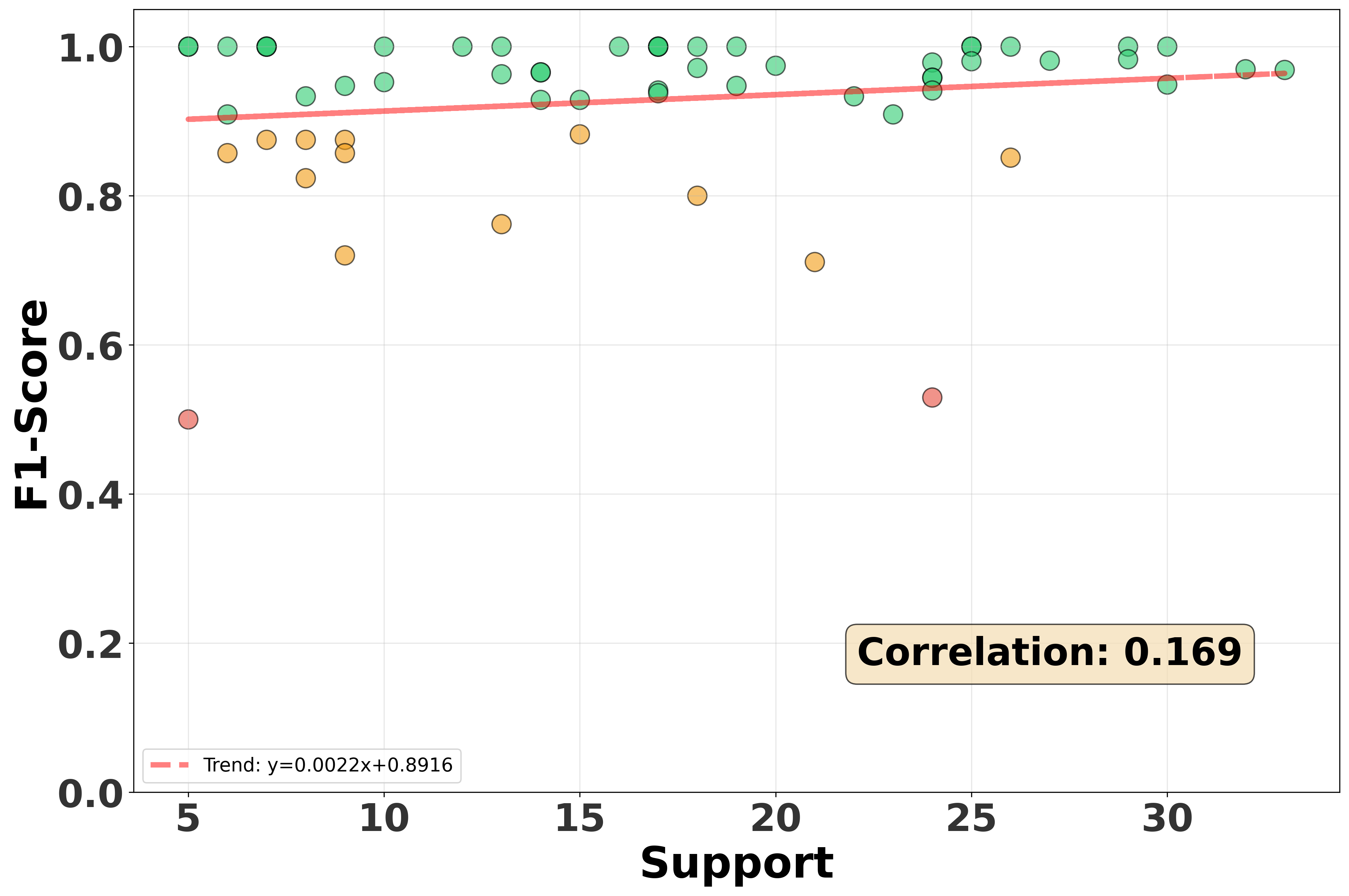}
    \caption{Insensitivity to class Imbalance}
    \label{fig:ins}
\vspace{-0.1in}
\end{figure}

\begin{table}[!htbp]
\centering
\begin{tabular}{l l c c}
\toprule
\textbf{Dataset} & \textbf{Annotator} & \textbf{Reliability} & \textbf{Accuracy} \\
\midrule
\multirow{7}{*}{Kinetics-400}
 & GPT-5-mini  & 0.964 & 0.933 \\
 & Gemini      & 0.944 & 0.936 \\
 & GPT-4o-mini & 0.889 & 0.847 \\
 & Pegasus-1.2 & 0.886 & 0.861 \\
 & Qwen-VL     & 0.741 & 0.718 \\
 & LLaVA-13b   & 0.645 & 0.642 \\
 & MoonDream2 & 0.065 & 0.058 \\
\midrule
\multirow{4}{*}{ImageNet-ReaL}
 & GPT-4o & 0.884 & 0.814 \\
 & GPT-4.1 Mini  & 0.874 & 0.759 \\
 & Grok 2   & 0.849 & 0.751 \\
 & GPT-4.1 Nano     & 0.427 & 0.396 \\
 \midrule
\multirow{4}{*}{Food-101}
 & Qwen3-VL    & 0.905 & 0.908 \\
 & Stella      & 0.826 & 0.770 \\
 & Nemesis  & 0.755 & 0.736 \\
 & InternVL & 0.644 & 0.634 \\
\midrule
\multirow{3}{*}{CUB-200}
 & Swin      & 0.909 & 0.876 \\
 & Gemini      & 0.827 & 0.803 \\
 & CLIP  & 0.727 & 0.714 \\
 & Random  & 0.586 & 0.600 \\
\bottomrule
\end{tabular}
\caption{Annotator Reliability Scores vs. Actual Accuracy Across Datasets}
\label{tab:annrel_combined}
 \vspace{-0.2in}
\end{table}

\vspace{-0.1in}
\subsection{Results}
{\bf Summary of Results.} 
The results demonstrate that {\tt AURA} is a generic and effective framework for data annotation and annotator reliability estimation, requiring neither domain-specific fine-tuning nor in-context learning. Across all experimental settings, {\tt AURA} consistently converges to a locally optimal solution and achieves higher annotation accuracy than any individual annotator. By leveraging complementary strengths of heterogeneous annotators while suppressing stochastic noise and systematic bias, {\tt AURA} produces robust aggregate labels. The inferred annotator reliabilities closely align with true accuracies under ground-truth evaluation, successfully distinguishing high and low-quality annotators. Compared to baseline, {\tt AURA}’s statistically grounded aggregation yields superior performance across nearly all metrics, as confirmed by ablation studies. Overall, {\tt AURA} exhibits stable convergence and strong resilience to class imbalance, making it well suited for a real-world annotation framework.


\begin{figure*}[!htbp]
\begin{center}
\begin{minipage}{0.28\textwidth}
\includegraphics[width=1\textwidth]{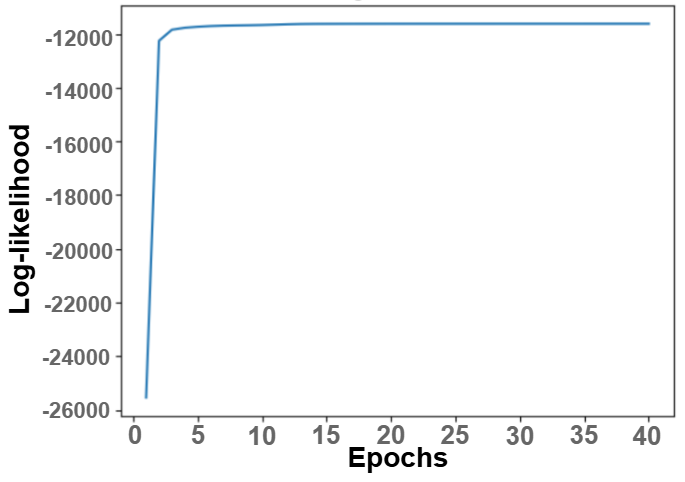}
\subcaption{Log-Likelihood change over epochs}\label{fig:convLL}
\end{minipage}
\begin{minipage}{0.29\textwidth}
\includegraphics[width=1\textwidth]{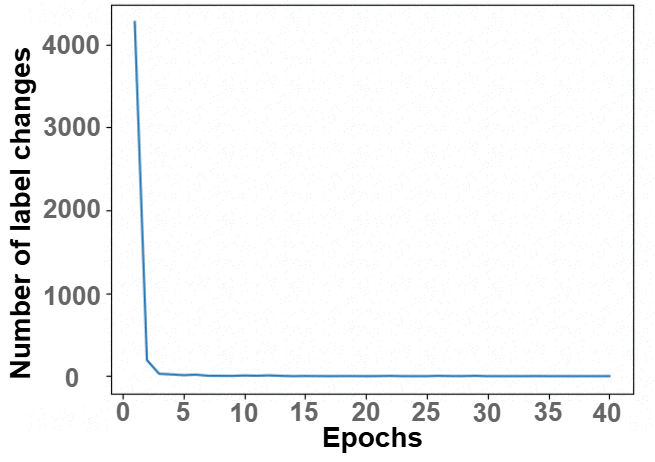}
\subcaption{Labels change over epochs}\label{fig:convlabel}
\end{minipage}
\begin{minipage}{0.27\textwidth}
\includegraphics[width=1\textwidth]{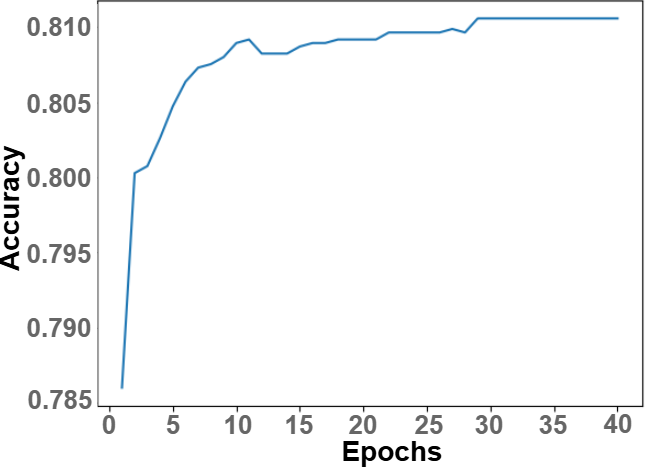}
\subcaption{Accuracy change over epochs}\label{fig:convacc}
\end{minipage}
\begin{minipage}{0.27\textwidth}
\includegraphics[width=1\textwidth]{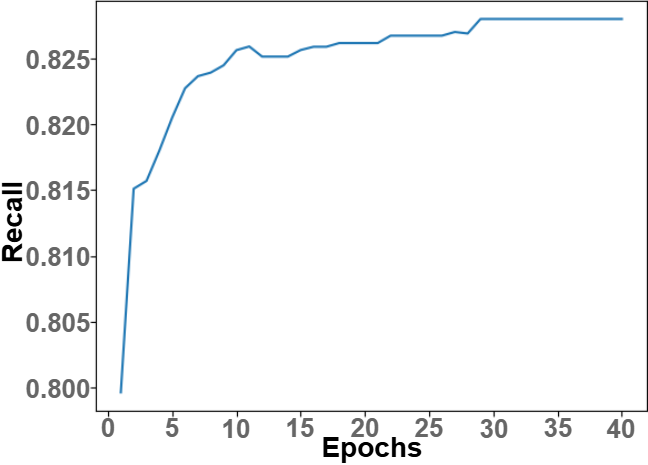}
\subcaption{Recall change over epochs}\label{fig:convrecall}
\end{minipage}
\begin{minipage}{0.27\textwidth}
\includegraphics[width=1\textwidth]{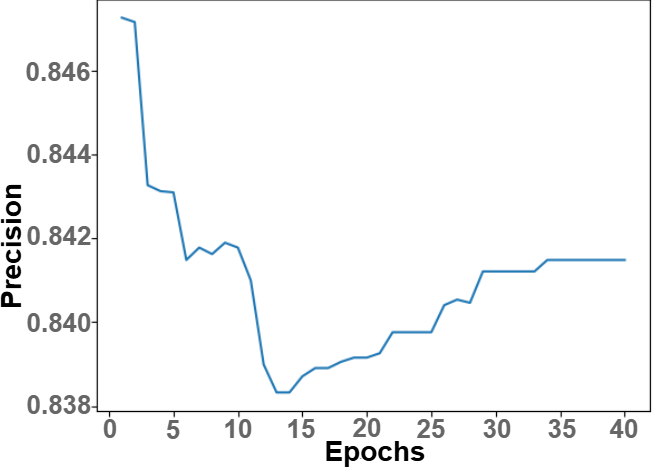}
\subcaption{Precision change over epochs}\label{fig:convprec}
\end{minipage}
\end{center}
\vspace{-0.1in}
\caption{\small Convergence of {\tt AURA}}\label{fig:conv}
\vspace{-0.2in}
\end{figure*}

\subsubsection{Performance Results}
{\tt AURA} achieves notably high accuracy across all datasets exceeding $80\%$ in all cases and surpassing $90\%$ on both Kinetics-400 and CUB-200 as observed in Table~\ref{tab:aura_all_metrics}, while consistently outperforming Majority Voting across all four benchmarks with performance gains ranging from $2.4\%$ (Kinetics-400) to $5.8\%$ (Food-101). To assess {\tt AURA}'s performance under reduced annotator quality, an ablation study was conducted on the Kinetics-400 dataset at the end of section~\ref{subsec:ablation}. 
{\tt AURA} exhibits a systematic precision-recall tradeoff, consistently achieving superior recall across all benchmarks, while showing marginally lower precision on three datasets. This observation reflects {\tt AURA}'s tendency towards higher sensitivity, often capturing a greater proportion of true positives while occasionally introducing false positives - consistent with framework's slight propensity for label overestimation. CUB-200 presents an exceptional case where both precision and recall were improved by {\tt AURA}, suggesting its reliability estimation mechanism is particularly effective for fine-grained annotation task with greater variance in annotator expertise. Despite occasional precision deficits, {\tt AURA} consistently outperforms Majority Voting in F1-scores demonstrating its enhanced recall performance yields a more favorable precision-recall balance and superior overall classification performance.

\noindent {\tt AURA} is insensitive to class imbalance as evidenced in Figure~\ref{fig:ins}. The low Pearson correlation coefficient of $0.169$ between F1-score and support suggests that the framework does not systematically favor majority classes or penalize minority ones in its aggregation process. Classes with lower F1-scores appear at different support levels indicating the inherent difficulty in annotating them rather than a lack of support issue. 

 \begin{figure*}[htbp]
    \centering
    \includegraphics[width=1\linewidth]{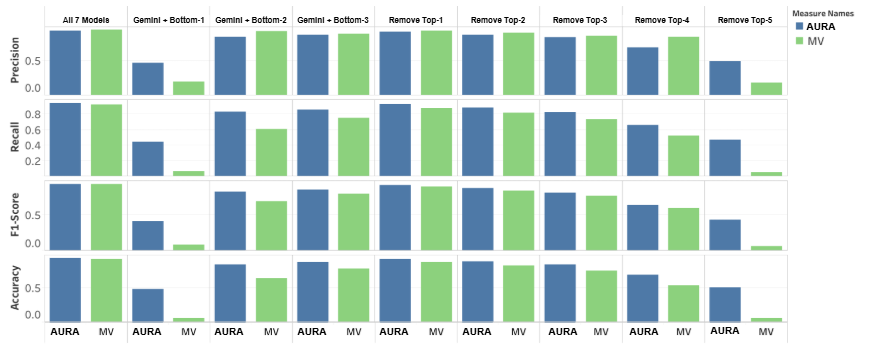}
    \caption{Ablation Study and Comparison with baseline}
    \label{fig:abl}
\vspace{-0.2in}
\end{figure*}

\subsubsection{Annotator Reliability vs. Accuracy of Annotators}\label{subsec:relvsaccuracy}
Table~\ref{tab:annrel_combined} comprehensively portrays {\tt AURA}'s robust reliability estimation capabilities across all 4 benchmark datasets and 18 heterogeneous annotators. The framework demonstrates consistent calibration performance, effectively retaining a strong rank-order link between estimated reliability and actual accuracy within each dataset, with typical discrepancies of 2-4\%. 
{\tt AURA} effectively identifies and appropriately downgrades poorly performing annotators, such as MoonDream2 (reliability: 0.065, accuracy: 0.058) in Kinetics-400, and GPT-4.1 Nano  (reliability: 0.427, accuracy: 0.396) in ImageNet-ReaL, while also closely approximating the expected random behavior for the Random annotator (reliability: 0.586, accuracy: 0.600). 
{\tt AURA}'s consistent performance across diverse annotator architectures including large language models (GPT, Gemini), specialized vision models (Swin, CLIP), and domain-specific systems (Pegasus-1.2, QwenVL), demonstrate its adaptability and independence from task-specific calibration or architectural assumptions.
{\tt AURA}'s reliability estimates serve as dependable proxies for actual annotator quality assessment in the absence of prior validation.

\subsubsection{Convergence of the Model}\label{subsec:converge}
The convergence behavior of {\tt AURA}, illustrated in Figure~\ref{fig:convLL}, is consistent across all datasets. The log-likelihood demonstrates monotonic increase until convergence to a local optimum. We also observe how the drastic change of labels in Figure~\ref{fig:convlabel} coincides with the increase in log-likelihood, indicating {\tt AURA} rapidly identifies optimal label assignments during early training phase. This increasing log-likelihood directly corresponds to improved annotation quality, as evidenced in Figures~\ref{fig:convacc} and~\ref{fig:convrecall}, where both accuracy and recall increase progressively across epochs, closely mirroring the log-likelihood trajectory before stabilizing upon convergence. Precision (figure~\ref{fig:convprec}) declines marginally, which is consistent with {\tt AURA}'s observed tendency toward label overestimation.

\subsubsection{Ablation Study}\label{subsec:ablation}

The ablation study conducted on the Kinetics-400 dataset in Figure~\ref{fig:abl} demonstrates  {\tt AURA}'s  remarkable robustness to low quality annotators maintaining substantially higher performance across most metrics in scenarios such as Gemini + Bottom-1/2/3 (in terms of accuracy, refer to Kinetics-400 annotators in Table~\ref{tab:annrel_combined}), where Majority Voting experiences severe degradation approaching near-zero performance. This resilience stems from {\tt AURA}'s reliability weighting mechanism which effectively mitigates the impact of poor annotators, whereas Majority Voting's equal treatment renders it vulnerable to quality variation. In practice, {\tt AURA} presents an extremely beneficial approach where a small number of costly, high-quality annotators with large number of cheap, low-quality alternatives can be utilized without compromising overall annotation quality. 

\noindent As top-performing annotators are progressively removed, {\tt AURA} consistently reigns superior in accuracy, recall and F1-scores, thereby showing its ability to balance the complementary expertise of the available annotators and extract maximum utility. Precision results corroborate {\tt AURA}'s overestimation tendency in almost every setting. 
In many real-world scenarios, the true quality of annotators is unknown or varies unpredictably over tasks. {\tt AURA} ensures the annotation system remains reliable even when some annotators perform poorly, eliminating the need for extensive pre-validation or continuous quality monitoring.

%% file: related.tex
\section{Related Work}\label{sec:related}
{\em Agentic AI frameworks.}
Several agentic AI frameworks combine multiple large language models to achieve complex behaviors. For example, LangChain~\cite{langchain} enables developers to build autonomous agents by integrating LLMs with external tools through a ReAct (Reasoning + Action) loop, where models iteratively reason, invoke tools, and incorporate feedback. Similarly, Microsoft AutoGen~\cite{autogen2024} supports the orchestration of LLM-powered agent workflows, offering both programmatic APIs and a no-code, pipeline-based interface, with strong support for event-driven agent interactions.

{\em Agentic AI for Data Annotation.} Recent work has demonstrated the promise of large language models (LLMs) as automated annotators. Several studies evaluate LLMs as standalone substitutes for human annotators, showing competitive or even superior performance on well-defined text annotation tasks such as event extraction and classification \cite{agentic1,agentic3}. Other efforts move beyond single-model assumptions by incorporating prompt diversity or in-context learning to capture variability in LLM behavior and to estimate annotator reliability \cite{agentic2,agentic4}. These approaches acknowledge that LLM outputs are not deterministic and that aggregation can improve label quality; however, they largely remain task-specific, heuristic in aggregation, or treat reliability implicitly rather than as a first-class modeling objective.

A complementary line of work explores agentic and multi-agent systems, where multiple LLM-based agents collaborate, reason, and refine outputs \cite{agentic5,agentic6,agentic7,agentic8}. While these systems emphasize autonomy, coordination, and reasoning efficiency, they are not designed explicitly for trustworthy annotation under uncertainty. 

{\em In contrast, \texttt{AURA} is generic, does not require any domain specific fine-tuning, and models heterogeneous AI annotators and aggregate their annotation while quantifying these models reliability in statistically grounded probabilistic fashion.}

%% file: conc.tex
\section{Conclusion}\label{sec:conc}
In this paper, we introduce {\tt AURA}, the first fully automated agentic AI framework for large-scale data annotation that relies exclusively on off-the-shelf AI models, without requiring any fine-tuning or in-context learning. {\tt AURA} builds on a classical probabilistic aggregation model to jointly infer latent true labels and the reliability of individual annotators. Extensive experiments demonstrate that {\tt AURA} consistently outperforms baselines across diverse datasets and settings, while exhibiting stable convergence behavior that reliably attains local optima. {\tt AURA} achieves substantially higher annotation accuracy than any individual annotator, establishing it as a dependable annotation framework in scenarios where annotator quality is unknown or heterogeneous. {\tt AURA} demonstrates robustness to class imbalance and maintains stable performance under skewed data distributions, making it well suited for real-world annotation pipelines. The estimated reliability coefficients closely track ground-truth annotator accuracies and successfully distinguish between high- and low-performing annotators. 

As an ongoing work, we are investigating how to incorporate heterogeneous costs in the AI models and design a framework that exhibits cost-performance trade-off.